\PassOptionsToPackage{table}{xcolor}
\documentclass[10pt,letterpaper]{article} \usepackage{preprint,times}

\usepackage{amsmath,amsfonts,bm}

\def\eqref#1{equation~\ref{#1}}

\def\1{\bm{1}}

\DeclareMathAlphabet{\mathsfit}{\encodingdefault}{\sfdefault}{m}{sl}
\SetMathAlphabet{\mathsfit}{bold}{\encodingdefault}{\sfdefault}{bx}{n}

\newcommand{\R}{\mathbb{R}}

\usepackage{float}
\usepackage[hidelinks]{hyperref}
\hypersetup{
  pdftitle={Spectral-MoE: Heterogeneous Per-Layer Expert Utilization in Dense-to-MoE Conversion},
  pdfauthor={},
  pdfsubject={},
  pdfkeywords={}
}
\ifdefined\pdfinfoomitdate\pdfinfoomitdate=1\fi
\ifdefined\pdftrailerid\pdftrailerid{}\fi
\ifdefined\pdfsuppressptexinfo\pdfsuppressptexinfo=15\fi
\usepackage{enumitem}
\usepackage{wrapfig}
\usepackage{zref-savepos,zref-abspage}
\usepackage{afterpage}
\usepackage{booktabs}
\usepackage{multirow}
\usepackage{longtable}
\usepackage{array}   \usepackage{graphicx}
\usepackage{tikz}
\usetikzlibrary{tikzmark}
\definecolor{grpband}{gray}{0.94}
\definecolor{specbandanalytic}{RGB}{196, 231, 248}
\definecolor{specbandlearned}{RGB}{254, 254, 203}

\newcommand{\nodata}{\textcolor{black!30}{--}}
\newcommand{\grouplab}[1]{\multirow{-3}{*}{$\vcenter{\hbox{\shortstack{\textbf{Spectral-MoE}\\#1}}}\,\left\{\rule{0pt}{16pt}\right.$}}

\DeclareMathOperator{\sg}{sg}
\DeclareMathOperator{\gelu}{GELU}

\DeclareMathOperator{\rms}{RMS}

\title{Spectral-MoE: Heterogeneous Per-Layer Expert Utilization in Dense-to-MoE Conversion}

\author{
  \begin{tabular*}{\textwidth}{@{\extracolsep{\fill}}ccc@{}}
    \textbf{Zachary Broveak} & \textbf{Christian Kubicka} & \textbf{Thierry Tambe}\\
    Stanford University & Stanford University & Stanford University\\
    {\small\ttfamily\href{mailto:zbrov@stanford.edu}{zbrov@stanford.edu}} &
    {\small\ttfamily\href{mailto:kubickce@stanford.edu}{kubickce@stanford.edu}} &
    {\small\ttfamily\href{mailto:ttambe@stanford.edu}{ttambe@stanford.edu}}
  \end{tabular*}
}
\date{}

\begin{document}

\maketitle

\begin{abstract}
Large language models impose substantial memory demands and arithmetic costs during inference. 
Dense-to-mixture-of-experts (MoE) conversion is a model optimization technique that aims to reduce these arithmetic costs.
Dense-to-MoE conversion partitions pretrained feed-forward network (FFN) neurons into experts and selects a subset for each token. 
Leading methods fall into two categories: analytic methods produce a converted model quickly using heuristics, while learning methods produce a model much more slowly by optimizing via gradients. 
Both approaches, however, fix the same active expert count, or “utilization,” at every layer.
We introduce Spectral-MoE, which learns to allocate a utilization budget heterogeneously across layers. 
Its name draws a metaphor between separating a dense FFN into experts and separating a light beam into its spectrum. 
Spectral-MoE consists of two distinct approaches: a static approach that uses a single per-layer scalar which is
shared across tokens, and a dynamic approach which emits a per-token schedule.
We demonstrate that both allocators improve performance over uniform on both the analytic-based and learning-based SOTA dense-to-MoE methods. 
At $25\%$ FFN utilization, dynamic allocation reduces WikiText-2 perplexity relative to uniform by $1.04$, $2.96$, and $2.71$ points on the learned base for Llama-2-7B, Llama-3-8B, and Qwen2.5-7B, respectively. 
The corresponding reductions on the analytic base are $2.45$, $6.26$, and $2.17$ points. On the learned base, dynamic allocation improves six-task mean accuracy over uniform by an average of $1.5$, $4.2$, and $3.6$ percentage points across the three models at FFN utilizations of $12.5\%$, $25\%$, and $50\%$, respectively. These results establish per-layer expert utilization as a novel
optimization axis in dense-to-MoE conversion.
\end{abstract}

\noindent\makebox[\linewidth][c]{\small\href{https://github.com/ZacharyBroveak/Spectral-MoE}{\raisebox{-0.2ex}{\includegraphics[height=1.8ex]{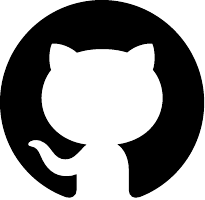}}\hspace{0.45em}\textbf{Code}}}

\section{Introduction}
\label{sec:intro}

\setlength{\intextsep}{5pt}
\begin{wrapfigure}{r}{0.5\linewidth}
  \centering
  \setlength{\abovecaptionskip}{4pt} \setlength{\belowcaptionskip}{0pt}
  \includegraphics[width=\linewidth]{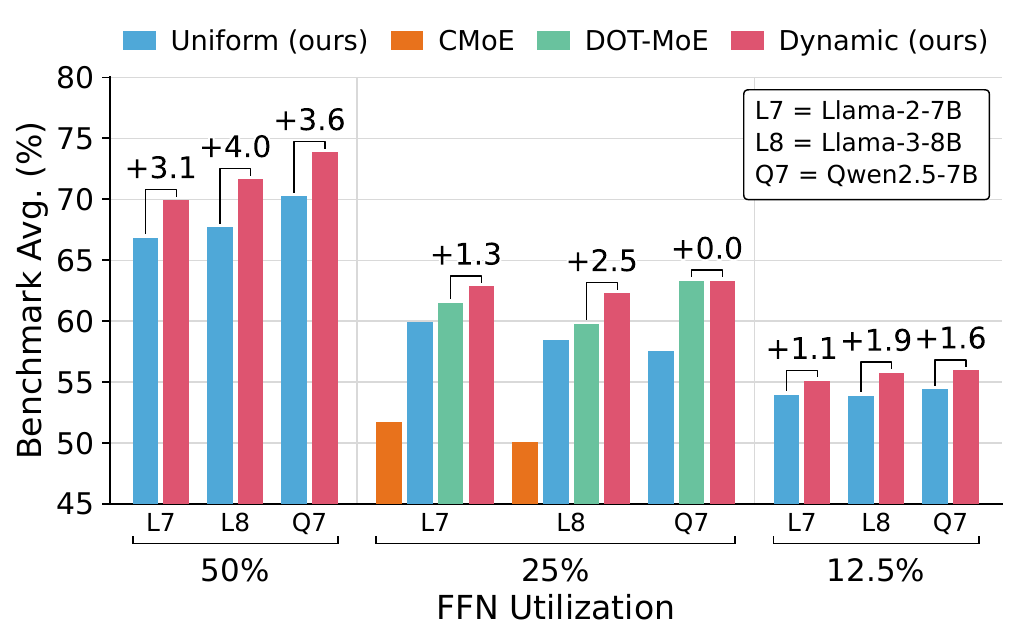}
  \caption{Benchmark accuracy versus utilization. Dynamic, heterogeneous allocation (learned) extends the frontier of dense-to-MoE performance.}
  \label{fig:fig1_bars}
\end{wrapfigure}

The scaling of large language models (LLMs) has produced remarkable capabilities in natural language understanding and generation \citep{brown2020fewshot}.
Realizing these capabilities, however, requires substantial computation
and memory bandwidth, with the latter becoming increasingly problematic for performance~\citep{gholami2024memorywall}.
Optimizing the transformer \citep{vaswani2017attention}, the engine powering
nearly all LLMs, has therefore become an extensive area of research. In this domain
there exist two primary axes along which compression research is performed:
quantization \citep{frantar2023gptq,nagel2020adaround,xiao2023smoothquant,lin2024awq}, and pruning \citep{frantar2023sparsegpt,sun2024wanda,han2015connections}.
A third technique, orthogonal to both quantization
and pruning, called dense-to-MoE conversion has also begun to emerge \citep{zhang2022moefication}.

A dense FFN evaluates every neuron for every token.
Dense-to-MoE conversion groups these neurons into experts and routes each token to a
subset, retaining pretrained weights while reducing active computation.
The fraction of experts active for a given token is the ``utilization'' and the assignment of neurons to expert groups is the neuron-to-expert ``partition''.
Existing methods differ in how they construct the partition
and the token-to-expert router
\citep{zhang2022moefication,qu2024llamamoev2,pei2025cmoe,bamba2026dotmoe, szatkowski2024exploitingactivationsparsitydense},
but all either use a uniform per-layer allocation or a floating utilization budget.
However, research on layer-dependent sensitivity to compression in quantization and pruning work
\citep{dong2019hawq,dong2020hawqv2,yin2024owl} motivates investigating
heterogeneous per-layer allocation as a potential optimization axis in dense-to-MoE conversion.

To this end, we present Spectral-MoE, which learns per-layer allocation under a fixed utilization budget via an allocator agnostic to the base dense-to-MoE conversion method. Our contributions are as follows:

\begin{enumerate}[leftmargin=1.7em, topsep=0.35em, itemsep=0.2em]
  \item \textbf{Heterogeneous per-layer allocation.} We introduce the first formulation of learnable per-layer allocation under a fixed per-token budget for dense-to-MoE conversion. This adds a third optimization axis to dense-to-MoE conversion alongside expert partitioning and routing.
  \item \textbf{Static and dynamic allocator tiers.} We develop a static allocator that learns one scalar per layer shared across tokens and a dynamic allocator that learns a per-token schedule. Both of these techniques are able to adhere to a user defined utilization budget.
  \item \textbf{Consistent performance gains.} We evaluate both static and dynamic allocators on the two leading approaches to dense-to-MoE conversion: analytic methods that convert the model quickly using heuristics \citep{pei2025cmoe}, and learning methods that optimize the model through slower gradient-based training \citep{bamba2026dotmoe}. Using our approach, we achieve SOTA mean accuracy at $25\%$ utilization and outperform uniform across budgets (Figure~\ref{fig:fig1_bars}).
\end{enumerate}

\section{Background and Related Work}
\label{sec:related}

\subsection{Heterogeneous Allocation}
\label{sec:related-alloc}

\paragraph{Per-Layer Allocation.}
Prior work on per-layer allocation typically uses pretrained MoEs,
with LTE \citep{zheng2024lte} as an exception (Section~\ref{sec:related-conversion}).
Alloc-MoE's Alloc-L uses dynamic programming over profiled layer sensitivities
\citep{liu2026allocmoe}, while LExI uses evolutionary search over per-layer
output-perturbation estimates \citep{chittyvenkata2025lexi}.
Both optimize surrogates that predict a schedule's accuracy degradation
as a sum of per-layer sensitivities.
Spectral-MoE instead learns per-layer utilization directly through gradients.

\paragraph{Per-Token Allocation.}
Prior work on per-token allocation uses confidence-based routing, null
experts, or budget redistribution to vary utilization across tokens.
Confidence-based routing sets utilization from the concentration of
routing scores \citep{huang2024dynamicmoe}, while AdaMoE introduces null
experts to vary utilization \citep{zeng2024adamoe}.
SeqTopK and Alloc-MoE's Alloc-T redistribute a layer's utilization budget
across tokens \citep{wen2026seqtopk,liu2026allocmoe}.
Spectral-MoE's dynamic allocator maps hidden states to per-token schedules that conserve a utilization budget across layers.

\subsection{Dense-to-MoE Conversion}
\label{sec:related-conversion}

\paragraph{Analytic Methods.}
Analytic methods quickly construct neuron-to-expert partitions from heuristics.
MoEfication clusters neuron input weights \citep{zhang2022moefication}, while
LLaMA-MoE-v2 uses activation and gradient importance
\citep{qu2024llamamoev2}.
CMoE groups neurons by their activation patterns through balanced clustering,
then uses a representative neuron from each cluster to construct the router
\citep{pei2025cmoe}.
Spectral-MoE's analytic base uses CMoE to construct the partition and router, recovers weights with LoRA as CMoE does, then trains the allocator with both frozen.

\paragraph{Learning Methods.}
Learning methods use slower gradient-based optimization.
LLaMA-MoE partitions FFN neurons into experts before continued pretraining
\citep{zhu2024llamamoe}.
LTE jointly trains the router and model weights with an efficiency penalty and
heterogeneous utilization follows from thresholding router scores
\citep{zheng2024lte}. 
DOT-MoE jointly learns the neuron-to-expert partition and token-to-expert router
through differentiable relaxations
\citep{bamba2026dotmoe}.
Spectral-MoE learns per-layer utilization in tandem with this joint
optimization, enforcing the same total utilization budget for every token.

\clearpage
\begin{figure}[!t]
  \centering
  \setlength{\abovecaptionskip}{4pt}
  \setlength{\belowcaptionskip}{0pt}
  \includegraphics[width=1.0\linewidth]{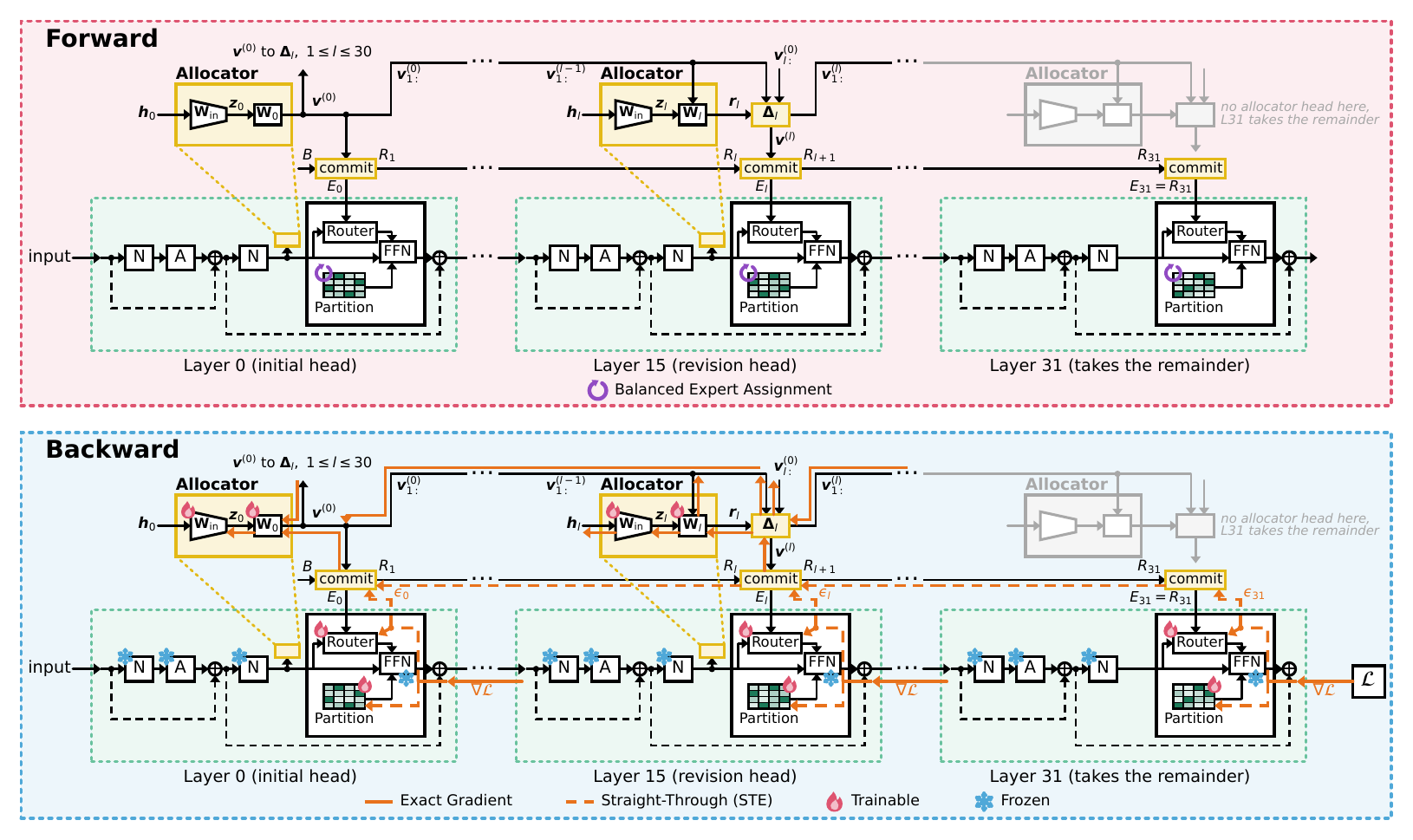}
  \caption{Spectral-MoE learned-base dynamic allocator architecture. During the forward pass, the allocator uses hidden states to initialize and revise per-token schedules, layers commit budget-constrained expert counts, and the final layer consumes the remainder. During the backward pass, loss gradients pass through selections and budget commitments via STEs to train only the allocator, routers, and partition affinities.}
  \label{fig:arch}
\end{figure}

\section{Methods}
\label{sec:method}

Spectral-MoE learns per-layer allocation under a budget derived from a user-specified target utilization.
This requires straight-through estimators (STE) \citep{bengio2013ste} and budget-preserving allocation.
We develop these components for static allocation, then extend them to a dynamic allocator that produces token-dependent schedules (Figure~\ref{fig:arch}).
Appendix~\ref{app:notation} summarizes the notation.

\subsection{Utilization Budget}
\label{sec:budget}

Consider $L$ FFN layers indexed by $l\in\{0,\ldots,L-1\}$.
Each layer contains $E_r$ routed and $E_s$ always-active shared experts, for $E_t=E_r+E_s$ experts in total.
Each expert contains $m$ neurons, giving FFN width $d_{\text{ffn}}=E_tm$.
We enforce a minimum number of routed experts $E_{\min}$ at each layer to ensure all experts are reachable.
We use $E_s=E_{\min}=1$.
For each token, the routed count $E_l$ is the number of routed experts selected at layer $l$, with $E_{\min}\le E_l\le E_r$.
The free count $f_l=E_l-E_{\min}$ is the number selected above the minimum, so per-layer FFN utilization is
\begin{equation*}
  u_l=\frac{E_s+E_l}{E_r+E_s}
      =\frac{E_s+E_{\min}+f_l}{E_r+E_s}.
\end{equation*}

The user specifies a nominal FFN utilization $u_{\mathrm{target}}$.
This determines the uniform initial routed count $\bar{E}$ and total routed budget $B$:
\begin{equation}
  \bar{E}=\left\lfloor u_{\mathrm{target}}(E_r+E_s)-E_s\right\rceil,
  \qquad B=L\bar{E},\qquad E_{\min}\le\bar{E}\le E_r.
  \label{eq:init}
\end{equation}
Here $\lfloor\cdot\rceil$ rounds to the nearest integer, with ties rounded up.
The target must yield a count within the stated bounds.
Both allocators enforce $\sum_l E_l=B$ for every token.

With mean free count $\bar{f}=\bar{E}-E_{\min}$, the re-distributable budget and realized utilization satisfy
\begin{equation*}
  \sum_l f_l=L\bar{f},\qquad
  u=\frac{1}{L}\sum_l u_l=\frac{\bar{E}+E_s}{E_r+E_s}.
\end{equation*}
Shared experts and minimum routed counts account for $L(E_s+E_{\min})$ expert activations per token, leaving $L\bar{f}$ to allocate.

Note that not all target utilizations are realizable.
For Llama-2-7B, $L=32$, $E_r=85$, and $E_s=1$, so a target of $0.25$ gives $\bar{E}=21$ and $B=672$.
The realized utilization is $22/86\approx25.58\%$ (Appendix~\ref{app:geometry}). Both allocators conserve this same integer budget.

\subsection{Static Per-Layer Allocation}
\label{sec:static}

Static allocation learns one continuous routed count $\epsilon_l$ per layer, shared across tokens.
We initialize all counts to $\bar{E}$ for a uniform allocation, but optimizer updates can violate the budget or count bounds.
After each update, we therefore adjust the counts to satisfy $\sum_l\epsilon_l=B$ and $\tfrac12\le\epsilon_l\le E_r-\tfrac12$.
Appendix~\ref{app:static} details how we enforce these bounds and resolve any budget excess or deficit.

\paragraph{Hard Expert Selection.}
Let $\Gamma_B$ denote largest-remainder rounding with integer-bound repair, producing the forward schedule $\bm{E}=\Gamma_B(\bm{\epsilon})$.
We floor each count, then add one to the $B-\sum_l\lfloor\epsilon_l\rfloor$ layers with the largest fractional parts.
If integer clipping changes the sum, we adjust eligible counts to restore $B$ (Appendix~\ref{app:static}).
Each router selects its top-$E_l$ experts per token, allowing expert identities to vary under the shared schedule.

\paragraph{Gradients Through Hard Selection.}
Let $\mathrm{rank}_l(t,e)\in\{0,\ldots,E_r-1\}$ rank expert $e$ for token $t$ at layer $l$, with zero assigned to the highest router score.
Treating ranks as constants during backpropagation, we define the soft inclusion
\begin{equation}
  \tilde{R}^{(l)}_{te}
  =\sigma\!\left(\frac{\epsilon_l-\mathrm{rank}_l(t,e)-\tfrac12}{T}\right),
  \label{eq:softincl}
\end{equation}
where $\sigma$ is the logistic sigmoid and $T>0$ the inclusion temperature (Appendix \ref{app:temp}). This makes soft inclusion most sensitive to $\epsilon_l$ for experts near the selection boundary.
Soft inclusions serve only to compute count gradients and need not sum to $\epsilon_l$.
The forward pass selects exactly $E_l$ routed experts at each layer, with the integer allocation rule enforcing $\sum_l E_l=B$.

For $n$ tokens, let $\mathbf{R}\in\{0,1\}^{n\times E_r}$ indicate routed selections, with $R_{te}=1$ when token $t$ selects expert $e$, and let $\mathbf{P}$ contain router softmax probabilities.
With $\sg$ denoting stop-gradient, we add the count-gradient term to the inherited router STE \citep{bamba2026dotmoe}:
\begin{equation}
  \mathbf{R}_{\mathrm{STE}}=\mathbf{R}
    +(\mathbf{P}-\sg[\mathbf{P}])
    +(\tilde{\mathbf{R}}^{(l)}-\sg[\tilde{\mathbf{R}}^{(l)}]).
  \label{eq:allocste}
\end{equation}
Its forward value is $\mathbf{R}$, while router and count gradients flow through $\mathbf{P}$ and $\tilde{\mathbf{R}}^{(l)}$, respectively.

\paragraph{Integration with the Conversion Base.}
The learned-base Spectral-MoE jointly trains partition affinities, router weights, and the allocator while keeping pretrained weights frozen.
The analytic-base Spectral-MoE first recovers weights with LoRA, then freezes weights, partition, and router while training the allocator.
Both bases retain an always-active shared expert, selected by activation rate on the analytic base and calibrated output contribution on the learned base.
The analytic base omits the router STE term in Equation 3. Inherited scalar routing coefficients scale expert outputs and count gradients. Training also evaluates the next-ranked expert, if available, with zero forward contribution. Its soft-inclusion gradient helps assess the benefit of selecting another expert.
Both learned and analytic bases use an objective which combines language-modeling cross-entropy with dense-teacher distillation.
The learned base also retains router regularization and load balancing.

\subsection{Dynamic Per-Layer Allocation}
\label{sec:pertoken}

Dynamic allocation revises each token's remaining schedule from the current hidden state.
A schedule predicted only at the first layer cannot respond to later hidden states.
A shared projection feeds an initial output head and subsequent revision heads.
The final layer has no revision head because it consumes the remainder.

\paragraph{Allocator Architecture.}
\label{sec:network}
Let $d_{\text{model}}$ be the residual-stream width and $q_l=L-l$ the number of uncommitted layers, including the current layer.
The shared projection $\mathbf{W}_{\mathrm{in}}$, initial head $\mathbf{W}_0$, and revision heads $\mathbf{W}_l$ have shapes
\begin{equation*}
  \mathbf{W}_{\mathrm{in}}\in\R^{L\times(d_{\text{model}}+1)},\qquad
  \mathbf{W}_0\in\R^{L\times L},\qquad
  \mathbf{W}_l\in\R^{q_l\times(L+q_l)}\quad(1\le l\le L-2),
\end{equation*}
where the additional input dimension of $\mathbf{W}_{\mathrm{in}}$ carries the log-RMS of the unnormalized FFN input.
Biases are omitted.
Including biases, the allocator adds $158{,}942$ parameters for $L=32$ and $d_{\text{model}}=4096$, negligible compared to the billion-scale models being converted.

For hidden state input $\bm{h}_l$, define its root-mean-square (RMS) normalized form $\bm{x}_l=\bm{h}_l/\rms(\bm{h}_l)$.
The schedule $\bm{v}^{(l)}\in\R^{q_l}$ contains logits for layers $l,\ldots,L-1$, with entry zero corresponding to the current layer.
Let $\bm{r}_l$ be a proposed revision and $\bm{\varDelta}_l$ the accumulated correction to the initial schedule.
The $\tanh$ operations bound each proposed logit revision and each logit's accumulated deviation from the initial schedule.
Square brackets concatenate features:
\begin{align*}
  \bm{z}_l&=\gelu\big(\mathbf{W}_{\mathrm{in}}[\bm{x}_l,\log\rms(\bm{h}_l)]\big),
  &\bm{v}^{(0)}&=\mathbf{W}_0\bm{z}_0,\\
  \bm{r}_l&=\tanh\big(\mathbf{W}_l[\bm{z}_l,\bm{v}^{(l-1)}_{1:}]\big),
  &\bm{\varDelta}_l&=\tanh\big((\bm{\varDelta}_{l-1})_{1:}+\bm{r}_l\big),\\
  \bm{v}^{(l)}&=\bm{v}^{(0)}_{l:}+\bm{\varDelta}_l,
  &\bm{\varDelta}_0&=\bm{0}.
\end{align*}
Concatenating the projected hidden state $\bm{z}_l$ with the remaining schedule $\bm{v}^{(l-1)}_{1:}$ lets each head correct the same hidden state differently depending on the allocation already proposed.
The log-RMS feature restores the scale removed by normalization.
Small nonzero head weights produce uniform integer counts on initialization while preserving gradients into the shared projection.

\paragraph{Exact Budget Conservation.}
\label{sec:conservation}
Let $S_l=\sum_{j<l}E_j$ be the routed budget committed at earlier layers $j$, $R_l=B-S_l$ its remainder, and $\pi_l=\operatorname{softmax}(\bm{v}^{(l)})_0$ the current layer's share.
After reserving $E_{\min}$ experts for each uncommitted layer, we assign share $\pi_l$ of the surplus to the current layer.
Let $a_l$ and $b_l$ denote the minimum and maximum feasible routed counts at the current layer, ensuring the remainder fits within later layers' count bounds.
We round and clip the continuous proposal $\epsilon_l$ to these bounds:
\begin{align*}
  \epsilon_l&=E_{\min}+\max(R_l-q_lE_{\min},0)\pi_l,\\
  a_l&=\max\{E_{\min},R_l-(q_l-1)E_r\},\qquad
  b_l=\min\{E_r,R_l-(q_l-1)E_{\min}\},\\
  E_l&=\operatorname{clip}_{[a_l,b_l]}\big(\lfloor\epsilon_l\rceil\big).
\end{align*}
Here $\lfloor\cdot\rceil$ rounds ties to even, and clipping preserves feasibility under any tie rule.

For a feasible initial budget, the bounds ensure $(q_l-1)E_{\min}\le R_l-E_l\le(q_l-1)E_r$ at every layer.
At the final layer, $q_l-1=0$, forcing $E_{L-1}=R_{L-1}$ and hence $\sum_l E_l=B$ for every token.

\paragraph{Gradients Through the Utilization Schedule.}
\label{sec:coupling}
The soft inclusion in Equation~\ref{eq:softincl} supplies gradients through the continuous proposal $\epsilon_l$ to both schedule shares and the remaining budget.
For positive surplus with $R_l$ fixed, the derivative with respect to schedule logit $v^{(l)}_j$ is
\begin{equation*}
  \frac{\partial\epsilon_l}{\partial v^{(l)}_j}
  =(\epsilon_l-E_{\min})
    \big(\delta_{0j}-\operatorname{softmax}(\bm{v}^{(l)})_j\big),
\end{equation*}
Here schedule index $j$ refers to layer $l+j$, and $\delta_{0j}$ is one for $j=0$ and zero otherwise.
Increasing a later layer's logit reduces the current layer's surplus share, coupling allocation decisions across the remaining layers.
We evaluate soft inclusion using $\epsilon_l$ clipped to $[a_l,b_l]$.
During backpropagation, we treat clipping as the identity, allowing gradients to reach the continuous proposal even when a bound is active.

Earlier integer commitments also affect later proposals by changing the remaining budget.
To preserve this gradient path, we compute $B-\widehat{S}_l$ using an STE over earlier layers $j$:
\begin{equation*}
  \widehat{S}_l=\sum_{j<l}\big(E_j+\epsilon_j-\sg[\epsilon_j]\big),\qquad
  \frac{\partial\epsilon_l}{\partial\widehat{S}_l}=-\pi_l.
\end{equation*}
The derivative assumes positive surplus and holds the current schedule logits constant.
The proposal terms cancel in the forward pass, giving $\widehat{S}_l=S_l$ and preserving the exact budget.
During backpropagation, $E_j$ is held constant and each summand passes gradients to the proposal $\epsilon_j$ before rounding or clipping.
Increasing the committed budget therefore reduces the current proposal at rate $\pi_l$, allowing later selections, including the final layer's, to influence earlier allocations.

\clearpage
\setlength{\textfloatsep}{10pt plus 2pt minus 2pt}
\begin{table}[!t]
  \centering
  \small
  \renewcommand{\arraystretch}{0.94}
  \setlength{\tabcolsep}{4.5pt}
\begin{tikzpicture}[remember picture]
\node[inner sep=0pt,outer sep=0pt] (canonicaltable) {\begin{tabular}{r@{\hspace{2.25pt}}lcccccccc}
  \toprule
  \multicolumn{2}{@{}l}{\multirow{2}{*}{\textbf{Method}}} & \multirow{2}{*}{\textbf{PPL ($\downarrow$)}} & \textbf{BoolQ} & \textbf{SciQ} & \textbf{PIQA} & \textbf{WinoG.} & \textbf{ARC-C} & \textbf{HellaS.} & \multicolumn{1}{c}{\multirow{2}{*}{\textbf{Avg.}}} \\
  \multicolumn{2}{l}{} & & {\scriptsize[32, acc]} & {\scriptsize[0, acc]} & {\scriptsize[0, acc]} & {\scriptsize[5, acc]} & {\scriptsize[25, acc-n]} & {\scriptsize[10, acc-n]} & \multicolumn{1}{c}{} \\
  \midrule
  \rowcolor{grpband}\multicolumn{10}{c}{Llama-2-7B} \\
  \multicolumn{2}{@{}l}{Dense} & 5.47 & 82.0 & 94.0 & 78.1 & 74.3 & 52.5 & 78.9 & \multicolumn{1}{c}{76.6} \\
  \addlinespace[4pt]
  \multicolumn{2}{l}{LLaMA-MoE$^\dagger$} & \tikzmark{canonical-0-analytic-top}\nodata & 37.8 & 20.0 & 49.7 & 50.1 & 25.8 & 26.2 & 34.9 \\
  \multicolumn{2}{l}{LLaMA-MoE-v2$^\dagger$} & \nodata & 51.3 & 67.0 & 56.6 & 52.9 & 25.7 & 35.1 & 48.1 \\
  \multicolumn{2}{l}{CMoE$^\dagger$} & \nodata & 55.0 & 77.5 & 57.1 & 54.1 & \textbf{27.6} & 38.8 & 51.7 \\
  \addlinespace[2pt]
  \rowcolor{specbandanalytic} & Uniform & 21.86 & 59.2 & 84.0 & 59.0 & 52.8 & 26.1 & 39.1 & 53.4 \\
  \rowcolor{specbandanalytic} & Static & 20.54 & 61.7 & \textbf{85.0} & 59.6 & \textbf{56.1} & 25.0 & 39.9 & \textbf{54.6} \\
  \rowcolor{specbandanalytic}\grouplab{\textbf{Analytic}} & Dynamic & \tikzmark{canonical-0-analytic-bottom}\textbf{19.41} & \textbf{61.8} & 83.1 & \textbf{60.2} & 54.8 & 26.5 & \textbf{41.5} & \textbf{54.6} \\
  \addlinespace[6pt]
  \multicolumn{2}{l}{DOT-MoE$^\ddagger$} & \tikzmark{canonical-0-learned-top}\nodata & 66.4 & 89.6 & 67.2 & 57.9 & 34.0 & 53.9 & 61.5 \\
  \addlinespace[2pt]
  \rowcolor{specbandlearned} & Uniform & 11.61 & 63.2 & 88.9 & 65.0 & 57.3 & 32.3 & 52.7 & 59.9 \\
  \rowcolor{specbandlearned} & Static & 11.50 & 64.8 & \textbf{89.8} & 65.3 & 58.4 & 33.6 & 52.8 & 60.8 \\
  \rowcolor{specbandlearned}\grouplab{\textbf{Learned}} & Dynamic & \tikzmark{canonical-0-learned-bottom}\textbf{10.57} & \textbf{67.1} & 88.8 & \textbf{68.4} & \textbf{60.5} & \textbf{35.8} & \textbf{56.5} & \textbf{62.8} \\
  \addlinespace[4pt]
  \midrule
  \rowcolor{grpband}\multicolumn{10}{c}{Llama-3-8B} \\
  \multicolumn{2}{@{}l}{Dense} & 6.14 & 83.2 & 96.2 & 79.6 & 77.3 & 58.3 & 82.1 & \multicolumn{1}{c}{79.4} \\
  \addlinespace[4pt]
  \multicolumn{2}{l}{LLaMA-MoE$^\dagger$} & \tikzmark{canonical-1-analytic-top}\nodata & 37.8 & 20.3 & 49.0 & 50.7 & 25.7 & 25.8 & 34.9 \\
  \multicolumn{2}{l}{CMoE$^\dagger$} & \nodata & 56.9 & 72.0 & 57.3 & 51.9 & \textbf{25.8} & 36.7 & 50.1 \\
  \addlinespace[2pt]
  \rowcolor{specbandanalytic} & Uniform & 45.24 & 60.9 & 82.2 & 58.9 & \textbf{52.5} & 23.7 & 36.8 & 52.5 \\
  \rowcolor{specbandanalytic} & Static & 40.48 & 61.9 & 82.1 & \textbf{59.7} & 51.1 & 24.1 & 36.5 & 52.6 \\
  \rowcolor{specbandanalytic}\grouplab{\textbf{Analytic}} & Dynamic & \tikzmark{canonical-1-analytic-bottom}\textbf{38.99} & \textbf{62.3} & \textbf{82.6} & 58.5 & \textbf{52.5} & 25.3 & \textbf{37.2} & \textbf{53.1} \\
  \addlinespace[6pt]
  \multicolumn{2}{l}{DOT-MoE$^\ddagger$} & \tikzmark{canonical-1-learned-top}\nodata & 62.3 & 90.9 & 65.0 & 56.0 & 34.1 & 50.1 & 59.8 \\
  \addlinespace[2pt]
  \rowcolor{specbandlearned} & Uniform & 19.40 & 59.1 & 87.2 & 64.5 & \textbf{58.2} & 32.3 & 49.3 & 58.5 \\
  \rowcolor{specbandlearned} & Static & 18.19 & 63.1 & 89.4 & 65.7 & 57.5 & 32.7 & 49.5 & 59.7 \\
  \rowcolor{specbandlearned}\grouplab{\textbf{Learned}} & Dynamic & \tikzmark{canonical-1-learned-bottom}\textbf{16.44} & \textbf{66.9} & \textbf{91.8} & \textbf{67.6} & \textbf{58.2} & \textbf{35.7} & \textbf{53.7} & \textbf{62.3} \\
  \addlinespace[4pt]
  \midrule
  \rowcolor{grpband}\multicolumn{10}{c}{Qwen2.5-7B} \\
  \multicolumn{2}{@{}l}{Dense} & 6.85 & 88.0 & 96.6 & 78.6 & 76.2 & 64.1 & 80.3 & \multicolumn{1}{c}{80.6} \\
  \addlinespace[4pt]
  \rowcolor{specbandanalytic} & Uniform & \tikzmark{canonical-2-analytic-top}47.07 & 59.4 & \textbf{78.3} & \textbf{58.7} & 53.0 & 25.6 & 35.9 & 51.8 \\
  \rowcolor{specbandanalytic} & Static & 45.61 & \textbf{60.3} & 76.8 & 58.2 & 52.7 & 25.3 & 36.0 & 51.5 \\
  \rowcolor{specbandanalytic}\grouplab{\textbf{Analytic}} & Dynamic & \tikzmark{canonical-2-analytic-bottom}\textbf{44.90} & 59.4 & 77.4 & 57.8 & \textbf{53.4} & \textbf{27.6} & \textbf{36.4} & \textbf{52.0} \\
  \addlinespace[6pt]
  \multicolumn{2}{l}{DOT-MoE$^\ddagger$} & \tikzmark{canonical-2-learned-top}\nodata & 64.3 & \textbf{93.9} & 68.5 & \textbf{58.7} & 39.6 & 55.0 & \textbf{63.3} \\
  \addlinespace[2pt]
  \rowcolor{specbandlearned} & Uniform & 16.92 & 43.7 & 89.4 & 66.9 & 58.0 & 35.4 & 51.9 & 57.6 \\
  \rowcolor{specbandlearned} & Static & 15.99 & \textbf{65.9} & 90.8 & 65.6 & 57.8 & 35.2 & 51.5 & 61.1 \\
  \rowcolor{specbandlearned}\grouplab{\textbf{Learned}} & Dynamic & \tikzmark{canonical-2-learned-bottom}\textbf{14.21} & 65.2 & 92.6 & \textbf{69.0} & 57.5 & \textbf{39.8} & \textbf{55.8} & \textbf{63.3} \\
  \addlinespace[4pt]
  \bottomrule
\end{tabular}
};
\begin{scope}[overlay]
\coordinate (canonical-0-analytic-t) at (pic cs:canonical-0-analytic-top);
\coordinate (canonical-0-analytic-b) at (pic cs:canonical-0-analytic-bottom);
\draw[draw=black,densely dotted,line width=0.75pt]
  ([yshift=9pt]canonicaltable.west |- canonical-0-analytic-t)
  rectangle ([yshift=-4pt]canonicaltable.east |- canonical-0-analytic-b);
\coordinate (canonical-0-learned-t) at (pic cs:canonical-0-learned-top);
\coordinate (canonical-0-learned-b) at (pic cs:canonical-0-learned-bottom);
\draw[draw=black,densely dotted,line width=0.75pt]
  ([yshift=9pt]canonicaltable.west |- canonical-0-learned-t)
  rectangle ([yshift=-4pt]canonicaltable.east |- canonical-0-learned-b);
\coordinate (canonical-1-analytic-t) at (pic cs:canonical-1-analytic-top);
\coordinate (canonical-1-analytic-b) at (pic cs:canonical-1-analytic-bottom);
\draw[draw=black,densely dotted,line width=0.75pt]
  ([yshift=9pt]canonicaltable.west |- canonical-1-analytic-t)
  rectangle ([yshift=-4pt]canonicaltable.east |- canonical-1-analytic-b);
\coordinate (canonical-1-learned-t) at (pic cs:canonical-1-learned-top);
\coordinate (canonical-1-learned-b) at (pic cs:canonical-1-learned-bottom);
\draw[draw=black,densely dotted,line width=0.75pt]
  ([yshift=9pt]canonicaltable.west |- canonical-1-learned-t)
  rectangle ([yshift=-4pt]canonicaltable.east |- canonical-1-learned-b);
\coordinate (canonical-2-learned-t) at (pic cs:canonical-2-learned-top);
\coordinate (canonical-2-learned-b) at (pic cs:canonical-2-learned-bottom);
\draw[draw=black,densely dotted,line width=0.75pt]
  ([yshift=9pt]canonicaltable.west |- canonical-2-learned-t)
  rectangle ([yshift=-4pt]canonicaltable.east |- canonical-2-learned-b);
\end{scope}
\end{tikzpicture}
  \caption{Dense-to-MoE conversion at 25\% FFN utilization. Headers give [shots, metric]. Downstream benchmarks use the evaluation harness from \citet{gao2024lmeval}. Analytic and Learned use CMoE and DOT-MoE bases, respectively. Uniform denotes our custom implementation of the bases.  Uniform, static, and dynamic runs are seed matched (Appendix~\ref{app:seeds}). $\dagger$: published \citet{pei2025cmoe}. $\ddagger$: published \citet{bamba2026dotmoe}.}
  \label{tab:tasks}
\end{table}

\section{Results}
\subsection{Experimental Settings}

\paragraph{Implementation.}
We implement Spectral-MoE using PyTorch \citep{paszke2019pytorch} and Hugging Face Transformers \citep{wolf2020transformers}, with learned-base conversion run on four NVIDIA H100 GPUs and analytic-base conversion run on one.
All converted models use 128 neurons per expert, following \citet{bamba2026dotmoe}, giving 86, 112, and 148 experts per layer for Llama-2-7B, Llama-3-8B, and Qwen2.5-7B, respectively.
On the analytic base, we construct the partition and router following CMoE \citep{pei2025cmoe}, recover weights with LoRA \citep{hu2022lora}, then freeze the recovered weights, partition, and router while training only the allocator.
On the learned base, we extend DOT-MoE's joint partition and router optimization \citep{bamba2026dotmoe} to train partition affinities, router weights, and allocator parameters in tandem while keeping pretrained weights frozen.
We optimize trainable parameters with AdamW \citep{loshchilov2019adamw}.
Appendix~\ref{app:protocol} details the full testing and training configurations.

\paragraph{Models and Evaluation.}
We evaluate Llama-2-7B \citep{touvron2023llama2}, Llama-3-8B \citep{grattafiori2024llama3}, and Qwen2.5-7B \citep{yang2024qwen25}.
We measure language-modeling performance using WikiText-2 test perplexity \citep{merity2017wikitext}.
Following the canonical benchmark selection of dense-to-MoE methods \citep{pei2025cmoe,bamba2026dotmoe}, we evaluate downstream benchmark performance using the harness provided by \citet{gao2024lmeval} on BoolQ \citep{clark2019boolq}, SciQ \citep{welbl2017sciq}, PIQA \citep{bisk2020piqa}, WinoGrande \citep{sakaguchi2020winogrande}, ARC-Challenge \citep{clark2018arc}, and HellaSwag \citep{zellers2019hellaswag}.

\paragraph{Baselines.}
CMoE \citep{pei2025cmoe} and DOT-MoE \citep{bamba2026dotmoe} represent the
current SOTA in analytic and learned conversion, respectively,
while LLaMA-MoE \citep{zhu2024llamamoe} and LLaMA-MoE-v2
\citep{qu2024llamamoev2} are representative of earlier work.
Given the substantial improvement of CMoE and DOT-MoE over previous work
(Table~\ref{tab:tasks}), we build Spectral-MoE on these two methods and
include the earlier methods as reference points.
We were unable to locate a public implementation of DOT-MoE during our
experiments. As a result, the reported uniform numbers are from our own implementation of DOT-MoE based on the details provided in the paper.
Given our baseline implementation falls short of the published accuracy, we report both sets of results.
Appendix~\ref{app:comparisons} describes our implementations of the conversion bases.

\begin{table}[!b]
  \centering
  \small
  \renewcommand{\arraystretch}{0.94}
  \setlength{\tabcolsep}{4pt}

  \newcommand{\sweeputil}[1]{\multirow{-6}{*}[1.5pt]{$#1\,\left\{
        \vcenter{\hbox{\rule{0pt}{\dimexpr5.64\baselineskip+3pt\relax}}}
      \right.$}}
  \centering
  \small
  \renewcommand{\arraystretch}{0.94}
  \setlength{\tabcolsep}{3.5pt}

  \edef\sweepbraceheight{\the\dimexpr5.64\baselineskip+3pt\relax}

  \def\sweepbrace{\multirow{-6}{*}[1.5pt]{$\smash{\left\{
        \vcenter{\hbox{\rule{0pt}{\sweepbraceheight}}}
      \right.}$}}

\begin{tabular}{c@{\hspace{2pt}}c@{\hspace{5pt}}llcccccccc}
  \toprule
  \multirow{2}{*}{\shortstack{\textbf{FFN}\\\textbf{Util.}}}
  &
  & \multirow{2}{*}{\textbf{Model}}
  & \multirow{2}{*}{\textbf{Allocator}}
  & \multirow{2}{*}{\textbf{PPL ($\downarrow$)}}
  & \textbf{BoolQ} & \textbf{SciQ} & \textbf{PIQA}
  & \textbf{WinoG.} & \textbf{ARC-C} & \textbf{HellaS.}
  & \multirow{2}{*}{\textbf{Avg.}} \\
  & & & & & {\scriptsize[32, acc]} & {\scriptsize[0, acc]}
  & {\scriptsize[0, acc]} & {\scriptsize[5, acc]}
  & {\scriptsize[25, acc-n]} & {\scriptsize[10, acc-n]} & \\
  \midrule

  & & Llama-2-7B & Uniform & 15.93 & 55.4 & 84.7 & \textbf{60.9} & \textbf{54.3} & \textbf{27.0} & 41.4 & 54.0 \\
  \rowcolor{specbandlearned}\cellcolor{white}
  & \cellcolor{white}
  & & Dynamic & \textbf{14.87} & \textbf{60.5} & \textbf{85.2} & 60.3 & 53.0 & 26.9 & \textbf{44.6} & \textbf{55.1} \\
  \addlinespace[1.5pt]
  & & Llama-3-8B & Uniform & 27.69 & 59.9 & 84.6 & 60.8 & 52.5 & 25.9 & 39.5 & 53.9 \\
  \rowcolor{specbandlearned}\cellcolor{white}
  & \cellcolor{white}
  & & Dynamic & \textbf{25.46} & \textbf{62.1} & \textbf{86.3} & \textbf{62.6} & \textbf{54.5} & \textbf{28.2} & \textbf{40.9} & \textbf{55.8} \\
  \addlinespace[1.5pt]
  & & Qwen2.5-7B & Uniform & 23.60 & 53.5 & 84.2 & \textbf{63.1} & \textbf{55.6} & 28.5 & 41.6 & 54.4 \\
  \rowcolor{specbandlearned}\cellcolor{white}
  \multirow{-6}{*}[1.5pt]{$0.125$}
  & \cellcolor{white}\sweepbrace
  & & Dynamic & \textbf{21.14} & \textbf{62.0} & \textbf{85.3} & 61.8 & 54.1 & \textbf{29.2} & \textbf{43.4} & \textbf{56.0} \\
  \midrule

  & & Llama-2-7B & Uniform & 11.61 & 63.2 & \textbf{88.9} & 65.0 & 57.3 & 32.3 & 52.7 & 59.9 \\
  \rowcolor{specbandlearned}\cellcolor{white}
  & \cellcolor{white}
  & & Dynamic & \textbf{10.57} & \textbf{67.1} & 88.8 & \textbf{68.4} & \textbf{60.5} & \textbf{35.8} & \textbf{56.5} & \textbf{62.8} \\
  \addlinespace[1.5pt]
  & & Llama-3-8B & Uniform & 19.40 & 59.1 & 87.2 & 64.5 & \textbf{58.2} & 32.3 & 49.3 & 58.5 \\
  \rowcolor{specbandlearned}\cellcolor{white}
  & \cellcolor{white}
  & & Dynamic & \textbf{16.44} & \textbf{66.9} & \textbf{91.8} & \textbf{67.6} & \textbf{58.2} & \textbf{35.7} & \textbf{53.7} & \textbf{62.3} \\
  \addlinespace[1.5pt]
  & & Qwen2.5-7B & Uniform & 16.92 & 43.7 & 89.4 & 66.9 & \textbf{58.0} & 35.4 & 51.9 & 57.6 \\
  \rowcolor{specbandlearned}\cellcolor{white}
  \multirow{-6}{*}[1.5pt]{$0.25$}
  & \cellcolor{white}\sweepbrace
  & & Dynamic & \textbf{14.21} & \textbf{65.2} & \textbf{92.6} & \textbf{69.0} & 57.5 & \textbf{39.8} & \textbf{55.8} & \textbf{63.3} \\
  \midrule

  & & Llama-2-7B & Uniform & 8.26 & 68.3 & 90.9 & 70.5 & 64.6 & 40.6 & 65.9 & 66.8 \\
  \rowcolor{specbandlearned}\cellcolor{white}
  & \cellcolor{white}
  & & Dynamic & \textbf{7.35} & \textbf{75.7} & \textbf{92.7} & \textbf{73.6} & \textbf{65.0} & \textbf{42.9} & \textbf{69.7} & \textbf{69.9} \\
  \addlinespace[1.5pt]
  & & Llama-3-8B & Uniform & 12.20 & 71.4 & 93.5 & 71.3 & 63.1 & 41.7 & 65.2 & 67.7 \\
  \rowcolor{specbandlearned}\cellcolor{white}
  & \cellcolor{white}
  & & Dynamic & \textbf{9.64} & \textbf{73.1} & \textbf{94.8} & \textbf{75.4} & \textbf{67.6} & \textbf{47.7} & \textbf{71.4} & \textbf{71.7} \\
  \addlinespace[1.5pt]
  & & Qwen2.5-7B & Uniform & 11.34 & 75.6 & 94.8 & 72.4 & 66.1 & 46.8 & 65.6 & 70.2 \\
  \rowcolor{specbandlearned}\cellcolor{white}
  \multirow{-6}{*}[1.5pt]{$0.50$}
  & \cellcolor{white}\sweepbrace
  & & Dynamic & \textbf{9.12} & \textbf{79.4} & \textbf{95.4} & \textbf{75.8} & \textbf{67.1} & \textbf{54.5} & \textbf{71.0} & \textbf{73.9} \\
  \bottomrule
\end{tabular}
  \caption{Spectral-MoE learned-base utilization sweep. Dynamic allocation improves perplexity and six-task mean accuracy over uniform across all tested models and utilizations. Metrics follow Table~\ref{tab:tasks}.}
  \label{tab:tasksweep}
\end{table}

\subsection{Benchmark Performance}

\paragraph{Comparison with Previous Works.}
Table~\ref{tab:tasks} compares Spectral-MoE with uniform allocation and
published methods at 25\% FFN utilization.
On the analytic base, dynamic allocation reduces perplexity by 2.17--6.26
points and improves six-task mean accuracy by 0.2--1.3 percentage points
over uniform, exceeding published CMoE mean accuracy by 3.0 points on
both Llama-2-7B and Llama-3-8B.
On the learned base, dynamic allocation reduces perplexity by 1.04--2.96
points and improves mean accuracy by 3.0--5.8 percentage points over uniform.
It exceeds published DOT-MoE mean accuracy by 1.3 and 2.5 points on
Llama-2-7B and Llama-3-8B, respectively, and matches Qwen2.5-7B at
63.3\% at the reported precision.
Both allocators lower perplexity across all models and bases, demonstrating
robust benefits across different partition and router construction techniques.

\paragraph{Utilization Sweep.}
\label{sec:sweep}
Table~\ref{tab:tasksweep} compares Spectral-MoE's dynamic allocator with
uniform allocation on the learned base at 12.5\%, 25\%, and 50\% FFN utilization.
Perplexity and six-task mean accuracy improve for every model--budget pair,
with accuracy gains averaging 1.5, 4.2, and 3.6 percentage points across
models, respectively.
Perplexity falls by 1.06--2.46, 1.04--2.96, and 0.91--2.55 points
at 12.5\%, 25\%, and 50\% utilization, respectively.
These gains demonstrate that the benefits of dynamic allocation extend across the tested utilization range and motivate systematic utilization sweeps in future dense-to-MoE work.

\begin{figure}[!t]
  \centering
  \begin{tikzpicture}
    \node[anchor=south west, inner sep=0] (plots) at (0,0)
      {\includegraphics[width=\linewidth]{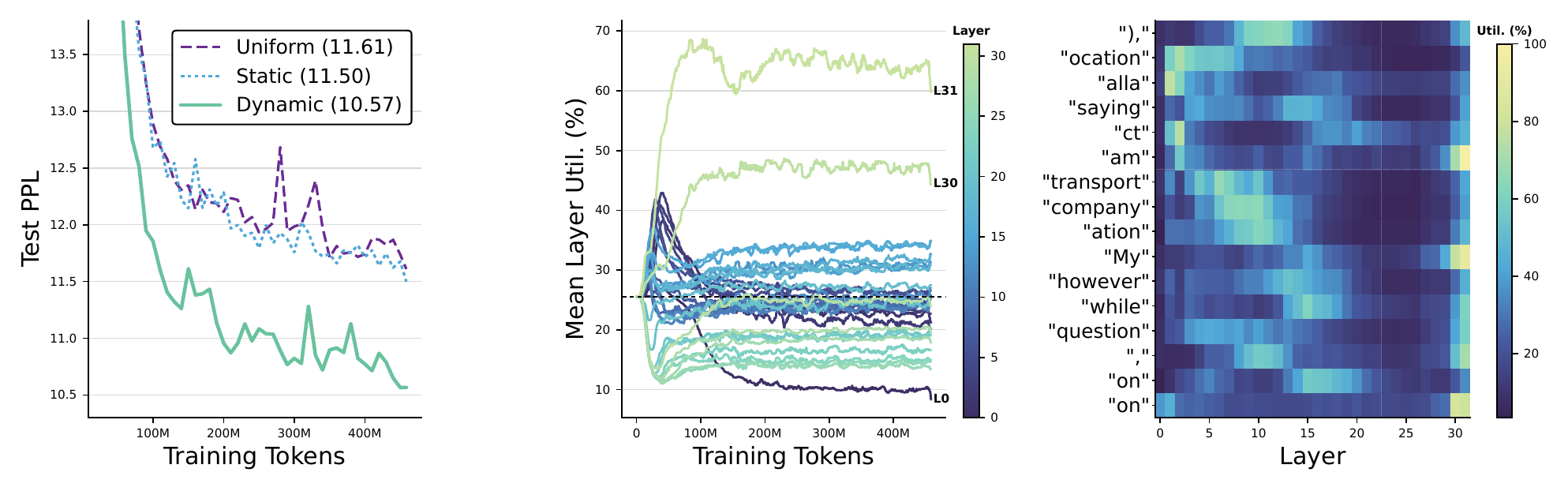}};

    \begin{scope}[x={(plots.south east)}, y={(plots.north west)}]
      \node[anchor=north, inner sep=0, font=\small] at (0.18,-0.02)
        {(a) Perplexity convergence};
      \node[anchor=north, inner sep=0, font=\small] at (0.50,-0.02)
        {(b) Per-layer convergence};
      \node[anchor=north, inner sep=0, font=\small] at (0.82,-0.02)
        {(c) Per-token schedules};
    \end{scope}
  \end{tikzpicture}
  \caption{Spectral-MoE on the Llama-2-7B learned base at a nominal 25\% FFN utilization.
  (a) WikiText-2 test perplexity.
  (b) Each layer's mean utilization under dynamic allocation. The dashed line marks the uniform start ($25.6\%$).
  (c) Final-step schedules of 16 sampled tokens. The two unique schedules for the token ``on'' showcase the allocator adapting to context.}
  \label{fig:heterogeneity}
\end{figure}

\subsection{Learned Heterogeneity}

\paragraph{Perplexity Convergence.}
Figure~\ref{fig:heterogeneity}(a) compares perplexity convergence on the
learned Llama-2-7B base at 25\% FFN utilization.
Dynamic allocation finishes at 10.57 perplexity, versus 11.61 for uniform
and 11.50 for static allocation.
Static allocation is initialized at uniform (Section~\ref{sec:static}), but ultimately learns a nonuniform schedule.
Its small final gain suggests a slightly improved allocation, although training
fluctuations prevent a firm conclusion.
Spectral-MoE's dynamic allocator, however, maintains a significant lead throughout.
Appendix~\ref{app:heterogeneity} provides the experimental specification and per-layer profiles across models.

\paragraph{Per-Layer Average Utilization.}
Figure~\ref{fig:heterogeneity}(b) tracks each layer's mean utilization
during dynamic allocation training.
The initially uniform schedule develops into a heterogeneous layer
profile whose averages stabilize late in training.
Together with the perplexity gains, this stabilization supports the assertion that
Spectral-MoE learns a useful allocation.
The resulting profile also offers a unique perspective on the relative
importance of each layer during dense-to-MoE conversion.

\paragraph{Per-Token Schedule Samples.}
Figure~\ref{fig:heterogeneity}(c) shows sampled per-token schedules from
the final training step.
The schedules vary substantially across tokens while preserving the
same total utilization budget.
Even positions sharing a token ID receive different allocations,
consistent with the allocator responding to changing hidden states.
Spectral-MoE's dynamic allocator can revise each token's remaining schedule,
combining stable layer averages with context-dependent allocation.
Appendix~\ref{app:heterogeneity} provides the experimental specification.

\subsection{Training Stability Considerations}
\label{sec:instability}

\begin{wrapfigure}{r}{0.48\linewidth}
  \vspace{-\intextsep}
  \centering
  \setlength{\abovecaptionskip}{4pt}
  \setlength{\belowcaptionskip}{0pt}
  \includegraphics[width=\linewidth]{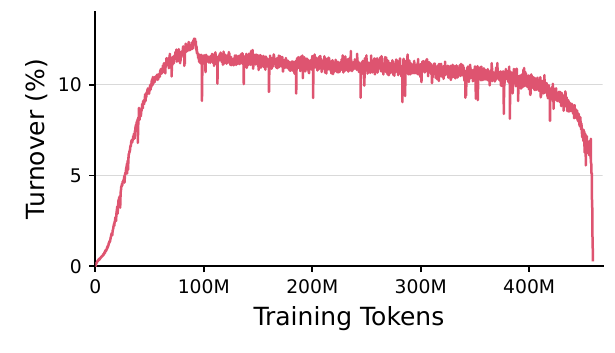}
  \caption{Fraction of routed neurons assigned to a different expert between consecutive training steps, measured on a sample learned-base, dynamic dense-to-MoE conversion.}
  \label{fig:partition-turnover}
\end{wrapfigure}

\paragraph{Neuron-to-Expert Assignment Turnover During Training.}
Figure~\ref{fig:partition-turnover} measures changes in neuron-to-expert
assignments during the dense-to-MoE conversion of a learned-base, dynamic Llama-2-7B model at $25\%$ FFN utilization.
Turnover is the fraction of routed neurons assigned to a different expert
between consecutive training steps, averaged across $32$ layers and excluding shared neurons.

\newpage

Turnover averages $8.92\%$ over the final $500$ steps but falls to $0.33\%$
at the final step.
The sharp drop at the end is consistent with the learning-rate decay toward zero, which eventually prevents updates to the partition affinities.
Prior to the sharp drop, however, the turnover rate appears to hold relatively constant despite the decreasing learning rate.
One issue sustained turnover introduces is that the router is forced to adapt to a constantly changing expert composition, potentially resulting in a partially adapted router at the end of joint training.  Despite this issue, learned dense-to-MoE conversion still performs quite well.

This observation motivates us to test whether additional router optimization after
freezing expert assignments boosts performance.
We leave the investigation of learning the partition itself in a more stable
fashion to future work.

\paragraph{Frozen Checkpoint Router Refitting.}
\label{sec:router-retraining}
To limit additional token use, we refit routers by closed-form ridge
regression~\citep{hoerl1970ridge}.
We pass $32{,}768$ training tokens through each frozen checkpoint from the end
of joint training, recording router inputs and computing an activation-based
importance proxy $I_{te}$ at each layer.
We then use these inputs to fit router weights and expert biases to
normalized regression labels $Y_{te}$, keeping all other parameters frozen.
For token $t$ and routed expert $e$,
\begin{equation}
  Y_{te}=\frac{I_{te}-\mu_t}{\max(s_t,10^{-6})},
  \qquad
  I_{te}=\sum_{j:M_{je}=1}|H_{tj}|\,
    \|\mathbf{W}_{\mathrm{down}}[:,j]\|_1,
  \label{eq:router-refit-labels}
\end{equation}
where $M_{je}=1$ assigns neuron $j$ to expert $e$, $H_{tj}$ is the neuron's FFN
activation, and $\mu_t,s_t$ are the mean and sample standard deviation of
$I_{te}$ across routed experts.

We compare each router's expert selections before and after refitting with
the expert ranking given by $I_{te}$.
All selections use the same pre-refit hidden states and expert counts.
``Overlap'' is the fraction of shared selections. ``Coverage'' is the selected
experts' summed importance divided by the reference total.
Both metrics improve for both checkpoints, but six-task mean accuracy changes
by $+1.0$ and $-0.4$ percentage points for seeds \textit{A} and \textit{B}
(Table~\ref{tab:router-retraining}). While the naive ridge regression router refitting yields mixed results, there may exist additional methods which improve accuracy with minimal token overhead. We leave the exploration of such approaches to future work. 

\setlength{\intextsep}{6pt}
\begin{table}[H]
  \centering
  \small
  \renewcommand{\arraystretch}{0.94}
  \setlength{\tabcolsep}{2.5pt}
  \begin{tabular}{@{}llrrrrrrrrr@{}}
    \toprule
    Seed & Router & Overlap & Coverage & BoolQ & SciQ & PIQA & WinoG. & ARC-C & HellaS. & Avg. \\
    \midrule
    \textit{A} & Original & 50.11 & 87.66 & 69.5 & 89.5 & 65.8 & 57.4 & 32.0 & 56.2 & 61.74 \\
    \rowcolor{specbandlearned} & Refit & 69.42 & 95.62 & 66.8 & 89.7 & 67.5 & 60.9 & 34.9 & 56.6 & 62.73 \\
    \addlinespace[2pt]
    \textit{B} & Original & 50.02 & 87.66 & 67.1 & 88.8 & 68.4 & 60.5 & 35.8 & 56.5 & 62.84 \\
    \rowcolor{specbandlearned} & Refit & 69.53 & 95.66 & 65.4 & 88.4 & 67.8 & 60.8 & 35.8 & 56.6 & 62.48 \\
    \bottomrule
  \end{tabular}
  \caption{Router refitting on learned dynamic Llama-2-7B at $25\%$ FFN utilization.
  Original and Refit denote checkpoint and ridge-regression routers.
  Entries are percentages; metrics follow Table~\ref{tab:tasks}.}
  \label{tab:router-retraining}
\end{table}

\section{Conclusion}
\label{sec:conclusion}
Spectral-MoE learns per-layer expert utilization during dense-to-MoE conversion under a user-specified budget.
Static allocation learns a schedule shared across tokens, while dynamic allocation revises each token's schedule and conserves the same total budget.
Dynamic allocation lowers perplexity on both conversion bases across all three models at nominal 25\% utilization, with learned-base perplexity and mean-accuracy gains extending to 12.5\% and 50\%.
It also exceeds published DOT-MoE mean accuracy on Llama-2-7B and Llama-3-8B and matches it on Qwen2.5-7B.
The learned schedules combine stable layer averages with context-dependent variation while preserving the same routed budget for every token.
Future work should examine whether stabilizing expert assignments improves the joint adaptation of routers and allocators.

\subsection*{AI Use Statement}
The main use of generative AI in this work was for code generation.  Specifically, we created a software specification manually and used generative AI to implement that specification.  The generated code was then reviewed manually to ensure correctness.  During the drafting and editing process of the manuscript generative AI was used to reword sentences, make certain paragraphs more concise, format equations, and identify related works for manual review.  The authors audited all AI generated artifacts and take full responsibility for the content of this work. 

\subsection*{Reproducibility Statement}
In order to ensure the reproducibility of our work we provide detailed training parameters and evaluation procedures in appendix \ref{app:protocol}.  The project repository and implementation release status are available at \url{https://github.com/ZacharyBroveak/Spectral-MoE}.

\clearpage
\appendix

\section{Notation}
\label{app:notation}

\begingroup
  \small
  \setlength{\tabcolsep}{0pt}
  \setlength{\LTcapwidth}{\linewidth}
  \setlength{\LTpre}{0pt}
  \setlength{\LTpost}{8pt}
  \renewcommand{\arraystretch}{1.08}
  \begin{longtable}{@{}
    >{\raggedright\arraybackslash}p{0.155\linewidth}
    @{\hspace{0.015\linewidth}}
    >{\raggedright\arraybackslash}p{0.305\linewidth}
    @{\hspace{0.03\linewidth}}
    >{\raggedright\arraybackslash}p{0.155\linewidth}
    @{\hspace{0.015\linewidth}}
    >{\raggedright\arraybackslash}p{0.315\linewidth}
    @{}}
  \caption{Notation used in the main text and appendices. Routed counts measure selected routed
  experts, whereas FFN utilization includes the shared expert. Bold italic symbols denote vectors
  and bold upright symbols denote matrices.}
  \label{tab:notation}\\
    \toprule
    Symbol & Definition & Symbol & Definition \\
    \midrule
    \endfirsthead
    \toprule
    Symbol & Definition & Symbol & Definition \\
    \midrule
    \endhead
    \bottomrule
    \endfoot

    $L$, $l$ & FFN layers and zero-based layer index. &
    $q_l$, $\pi_l$ & Uncommitted layers $L-l$, and the current layer's softmax share. \\

    $n$, $t$, $e$ & Token count, token index, and routed-expert index. &
    $\bm{v}^{(l)}$, $\bm{r}_l$, $\bm{\varDelta}_l$ & Vector of schedule logits, proposed
    revisions, and accumulated corrections in $\R^{q_l}$. \\

    $m$, $d_{\text{ffn}}$, $d_{\text{model}}$ & Neurons per expert, FFN width, and
    residual-stream width. &
    $S_l$, $R_l$, $\widehat{S}_l$ & Committed budget, remainder $B-S_l$, and the STE for
    $S_l$. \\

    $E_t$, $E_r$, $E_s$ & Total, routed, and shared experts per layer, with $E_t=E_r+E_s$. &
    $a_l$, $b_l$ & Feasible integer bounds on the current commitment. \\

    $E_{\min}$, $E_l$ & Minimum routed count and selected routed count at layer $l$. &
    $\bm{h}_l$, $\bm{x}_l$, $\bm{z}_l$ & FFN input, its $\rms$-normalized vector, and the allocator's shared-projection output. \\

    $f_l$, $\bar{f}$ & Free count $E_l-E_{\min}$ and its layer mean. &
    $\mathbf{W}_{\mathrm{in}}$, $\mathbf{W}_0$, $\mathbf{W}_l$ & Matrices for the shared projection,
    initial output head, and revision output head. \\

    $B$, $\bar{E}$ & Per-token routed budget and the uniform count $B/L$. &
    $\mathbf{H}$, $H_{tj}$ & FFN activation matrix and its scalar entry for token $t$, neuron $j$. \\

    $u_{\mathrm{target}}$, $u$, $u_l$ & User-specified utilization, realized mean, and per-layer utilization. &
    $\mathbf{W}_{\mathrm{down}}$ & FFN output projection matrix. \\

    $\epsilon_l$, $\bm{\epsilon}$ & Scalar continuous count or proposal for layer $l$, and the static count vector. &
    $\mathbf{M}$, $M_{je}$ & Hard neuron-to-expert partition matrix. $M_{je}=1$ assigns neuron $j$ to
    expert $e$. \\

    $\bm{E}$, $\Gamma_B$ & Integer count vector and largest-remainder rounding operator with bound repair. &
    $\mathbf{A}$ & Soft neuron-to-expert partition affinity matrix. \\

    $T$, $\sigma$ & Inclusion temperature and the logistic sigmoid. &
    $\mathbf{W}_r$, $\tau$ & Router weight matrix and partition transport temperature. \\

    $\mathrm{rank}_l(t,e)$ & Rank of expert $e$ by token $t$'s router score at layer $l$. &
    $\mathcal{L}$, $\nabla\mathcal{L}$ & Training objective and its gradient. \\

    $\mathbf{R}$, $\tilde{\mathbf{R}}^{(l)}$, $\mathbf{P}$ & Matrices of hard selection, soft inclusion,
    and router probabilities. &
    $p_{\mathrm{S}}$ & Student next-token distribution
    (Appendix~\ref{app:perplexity}). \\

    $\sg[\cdot]$ & Stop-gradient operation. &
    $I_{te}$ & Activation-based expert importance
    (Section~\ref{sec:router-retraining}). \\

    $\bm{g}$ & Objective gradient with respect to the static count vector
    (Appendix~\ref{app:static}). &
    $\mathcal{S}_{l,t}$, $\mathcal{O}_{l,t}$ & Selected and reference expert sets
    (Section~\ref{sec:router-retraining}). \\

    $x_i$, $x_{<i}$ & Target token and preceding context within its block. &
    $\mathbf{R}_{\mathrm{STE}}$ & STE for hard routed selection, used for router and
    count gradients. \\

    $\delta_{0j}$ & Kronecker delta: one when $j=0$, zero otherwise. & & \\
  \end{longtable}
\endgroup

\section{Methods Implementation Details}

\subsection{Real-Model Utilization Numbers}
\label{app:geometry}

Equation~\ref{eq:init} rounds
$u_{\mathrm{target}}(E_r+E_s)-E_s$ to an integer, restricting realized
utilization to multiples of $1/E_t$.
A requested utilization is exact when it is such a multiple and is otherwise
rounded to the nearest realizable value.
Llama-3-8B has $E_t=112$, making all three reported tiers exact.
Llama-2-7B and Qwen2.5-7B have $86$ and $148$ experts, respectively. They
realize one and two tiers exactly, with the remaining tiers within
$0.6$ percentage points of the requested utilization.

Table~\ref{tab:geometry} reports the geometry, realized utilization, and
routed budget $B=L\bar{E}$ for every model and tier.
Within each model and tier, all allocation methods hold $B$ fixed and
activate the same total number of FFN neurons for every token.
The per-layer average is $(\bar{E}+E_s)m$, or $2{,}816$ neurons on
Llama-2-7B at the nominal $25\%$ tier.
Exact agreement with the requested utilization is unnecessary for comparing
allocation methods, provided their realized budgets are matched.

\begin{table}[htb]
  \centering
  \small
  \setlength{\tabcolsep}{5.5pt}
  \begin{tabular}{@{}lrrrrrrr@{}}
    \toprule
    Model & $L$ & $E_t$ & $d_{\text{ffn}}$ & Nominal $u$ & $\bar{E}$ & Realized $u$ & $B$ \\
    \midrule
    \multirow{3}{*}{Llama-2-7B} & \multirow{3}{*}{32} & \multirow{3}{*}{86} & \multirow{3}{*}{11,008}
      & $12.5\%$  & 10 & $12.79\%$ & 320 \\
    & & & & $25\%$    & 21 & $25.58\%$ & 672 \\
    & & & & $50\%$    & 42 & $50.00\%$ & 1,344 \\
    \midrule
    \multirow{3}{*}{Llama-3-8B} & \multirow{3}{*}{32} & \multirow{3}{*}{112} & \multirow{3}{*}{14,336}
      & $12.5\%$  & 13 & $12.50\%$ & 416 \\
    & & & & $25\%$    & 27 & $25.00\%$ & 864 \\
    & & & & $50\%$    & 55 & $50.00\%$ & 1,760 \\
    \midrule
    \multirow{3}{*}{Qwen2.5-7B} & \multirow{3}{*}{28} & \multirow{3}{*}{148} & \multirow{3}{*}{18,944}
      & $12.5\%$  & 18 & $12.84\%$ & 504 \\
    & & & & $25\%$    & 36 & $25.00\%$ & 1,008 \\
    & & & & $50\%$    & 73 & $50.00\%$ & 2,044 \\
    \bottomrule
  \end{tabular}
  \caption{Expert geometry, the uniform routed count derived from each requested utilization, the
  utilization it realizes, and the conserved routed budget. All models use $m=128$ neurons per
  expert, one shared expert per layer, and $E_{\min}=1$.}
  \label{tab:geometry}
\end{table}

\subsection{Enforcing Bounds}
\label{app:static}

\paragraph{Utilization Update Stability.}
Static training constrains the continuous counts before and after each
optimizer step. Before the step, we center the count gradient,
$\bm{g}\leftarrow\bm{g}-(\bm{1}^{\top}\bm{g}/L)\bm{1}$,
so that the gradient is tangent to the budget hyperplane
$\{\bm{1}^{\top}\bm{\epsilon}=B\}$.
AdamW rescales each coordinate separately, so the resulting update can still
change the sum. After the step, we therefore restore the sum by projecting:
\begin{equation*}
  \bm{\epsilon}\leftarrow\bm{\epsilon}
    -\frac{\bm{1}^{\top}\bm{\epsilon}-B}{L}\bm{1},
\end{equation*}
which removes drift introduced by adaptive scaling, momentum, and finite
precision.

\paragraph{Enforcing Count Bounds.}
Centering the count gradient and restoring the sum does not enforce
$\tfrac12\le\epsilon_l\le E_r-\tfrac12$.  To mitigate this, violating coordinates are clamped and the projection over the remaining free coordinates is repeated.
The bounded fixed-sum set is acceptable when
$L/2\le B\le L(E_r-\tfrac12)$.  This formula holds for every budget in
Table~\ref{tab:geometry}.

\paragraph{Integer Repair.}
Flooring the continuous counts removes
$B-\sum_l\lfloor\epsilon_l\rfloor$ units. These units are assigned back to the layers with the largest fractional parts.  This process restores $\sum_l E_l=B$ and gives $|E_l-\epsilon_l|<1$ before bound repair.
Integer clipping can change the sum, so a seeded repair adds or removes one
expert at an eligible layer until the budget is restored.  Such a layer exists
whenever $LE_{\min}\le B\le LE_r$.
Continuous counts persist between steps, allowing sub-integer updates to
accumulate before the hard schedule changes.

\section{Experimental Protocol}
\label{app:protocol}

\subsection{Perplexity}
\label{app:perplexity}

\paragraph{Protocol.}
We measure WikiText-2 test perplexity \citep{merity2017wikitext} using each
model's tokenizer and complete, nonoverlapping blocks of $2{,}048$ tokens:
\begin{equation*}
  \mathrm{PPL}=\exp\!\left(
    -\frac{1}{\sum_{b}|b|}\sum_{b}\sum_{i\in b}
      \log p_{\mathrm{S}}(x_i\mid x_{<i})\right),
\end{equation*}
Here $b$ contains the scored next-token targets of a complete block, $i$ indexes these targets, and
$x_{<i}$ contains context only from that block. Preprocessing retains empty
rows and special tokens, and the trailing partial block is discarded.
The Llama-2-7B stream contains $341{,}469$ tokens. Its $166$ complete
blocks contain $339{,}968$ input tokens and $339{,}802$ scored targets.
Every final perplexity in this paper is the corresponding run's end-of-training
evaluation under this protocol, while training curves and step-$0$ values are
labeled where they appear.

\paragraph{Data Separation.}
All reported perplexities use the WikiText-2 test split.

\subsection{Benchmark Evaluation}

\paragraph{Benchmarks.}
Evaluation uses the harness provided by \citep{gao2024lmeval} with the
settings in Table~\ref{tab:downstream}. The six-task mean weights tasks
equally, and reported cross-model gains average the three model-specific gains
equally. Averages and differences are computed from unrounded scores, then
rounded for display.

\begin{table}[!htb]
  \centering
  \small
  \setlength{\tabcolsep}{6pt}
  \begin{tabular}{@{}lrlrl@{}}
    \toprule
    Task & Shots & Metric & Examples & Split \\
    \midrule
    BoolQ \citep{clark2019boolq} & 32 & accuracy & 3{,}270 & validation \\
    SciQ \citep{welbl2017sciq} & 0 & accuracy & 1{,}000 & test \\
    PIQA \citep{bisk2020piqa} & 0 & accuracy & 1{,}838 & validation \\
    WinoGrande \citep{sakaguchi2020winogrande} & 5 & accuracy & 1{,}267 & validation \\
    ARC-Challenge \citep{clark2018arc} & 25 & length-normalized accuracy & 1{,}172 & test \\
    HellaSwag \citep{zellers2019hellaswag} & 10 & length-normalized accuracy & 10{,}042 & validation \\
    \bottomrule
  \end{tabular}
  \caption{Evaluation settings, identical for every model and allocation method.}
  \label{tab:downstream}
\end{table}

\paragraph{Data Separation.}
Training and calibration use Dolmino \citep{olmo2025dolmino}.
Evaluation uses the WikiText-2 test split and the benchmarks in
Table~\ref{tab:downstream}.

\subsection{Training Configurations}

\paragraph{Learned Base}
Each learned-base run processes
$3{,}500\times64\times2{,}048=458{,}752{,}000$ Dolmino tokens from one
tokenized cache. Training uses bfloat16 on four H100 GPUs with sharded dense
weights and a shared data seed. The runs in Table~\ref{tab:tasks} use
initialization seed \textit{B}, and Table~\ref{tab:seeds} lists the others.
Partition affinities and router weights use a peak learning rate of
$5\times10^{-4}$ and weight decay $10^{-4}$.
Allocator parameters use a separate optimizer group with zero weight decay,
since decay on continuous counts would bias them toward zero routed experts.

\paragraph{Analytic Base.}  The base for the analytic approach is CMoE \citep{pei2025cmoe}.  Specifically, balanced $k$-means
\citep{lloyd1982,malinen2014balanced} with a linear-assignment step is used 
\citep{jonker1987lap} on $16{,}384$ Dolmino calibration tokens.  The
constructed Llama-2-7B model has perplexity $735.85$.
Following expert assignment, weight recovery trains LoRA adapters \citep{hu2022lora} of
rank $8$, scaling $\alpha=32$, and dropout $0.1$ on all attention and MLP
projections using $2{,}048$ samples of $2{,}048$ tokens.
The final perplexity is $21.86$.

The allocator then trains for $1{,}024$ steps over the same token budget,
with recovered weights, partition, and router frozen.
The analytic uniform baseline is the recovered checkpoint before allocator
training and therefore has zero allocator-training steps.
A comprehensive list of training parameters can be seen in Table~\ref{tab:hyper}.

\begin{table}[htb]
  \centering
  \small
  \setlength{\tabcolsep}{4pt}
  \begin{tabular}{@{}
    >{\raggedright\arraybackslash}p{0.28\linewidth}
    >{\raggedright\arraybackslash}p{0.35\linewidth}
    >{\raggedright\arraybackslash}p{0.29\linewidth}@{}}
    \toprule
    & Learned base & Analytic base \\
    \midrule
    Trained tensors & $\mathbf{A}$, $\mathbf{W}_r$, allocator & allocator only \\
    Frozen & pretrained FFN weights & weights, partition, router \\
    Steps / tokens & 3,500 / 458.8M & 1,024 / 4.19M \\
    Preceding stage & \nodata & LoRA recovery, 4.19M tokens \\
    $\mathbf{A}$, $\mathbf{W}_r$ peak rate & $5\times10^{-4}$ & \nodata \\
    Static $\bm{\epsilon}$ peak rate & $5\times10^{-3}\bar{E}/10$ & $0.875\,\bar{E}/512$ \\
    Dynamic network peak rate & $5\times10^{-5}\bar{E}/10$ & $5\times10^{-5}\bar{E}/10$ \\
    Inclusion temperature $T$ & 1.5 static, 1.0 dynamic & 1.0 \\
    Transport temperature $\tau$ & $1.0\!\rightarrow\!0.1$ over warmup, then flat & \nodata \\
    Optimizer & AdamW & AdamW \\
    Rate schedule & warmup over the first fifth, cosine to zero & as learned \\
    Evaluation / checkpoint cadence & every 10M / 50M tokens & end of run \\
    \bottomrule
  \end{tabular}
  \caption{Training configurations. The analytic static rate is set from the
  step count, $0.875\bar{E}$ spread over half the run, rather than from the learned base's constant.}
  \label{tab:hyper}
\end{table}

\subsection{Existing Method Implementations}
\label{app:comparisons}

The analytic base uses CMoE's published partition and router construction
\citep{pei2025cmoe}, unmodified.  The learned base, however, reimplements DOT-MoE due to a lack of available public code at the time of writing.  One modification was made to the original DOT-MoE work which was the inclusion of a shared expert for the purposes of training stability.

\begin{samepage}
\section{Run-to-Run Variation}
\label{app:ablations}
\label{app:seeds}

Dense-to-MoE conversion is sensitive to initialization meaning identical conversion
procedures can yield different perplexities and benchmark scores across seeds.
Estimating this variation across models, budgets, and allocation methods requires
many repeated experiments, whose computational cost is prohibitive.
Despite this cost, we did run a small number of different seeds in order to provide some indication of this variation.
Table~\ref{tab:seeds} reports representative statistics from the learned Llama-2-7B base training run at $25\%$ FFN utilization.\par
\end{samepage}

\begin{table}[H]
  \centering
  \small
  \setlength{\tabcolsep}{5.5pt}
  \begin{tabular}{@{}llrrr@{}}
    \toprule
    Allocation & Initialization & Step-0 PPL & Final PPL & Six-task avg. (\%) \\
    \midrule
    Dynamic & seed \textit{C} & 1115.01 & 10.72 & 62.2 \\
     & seed \textit{A} & 1279.96 & 10.59 & 61.7 \\
     & seed \textit{B} (Table~\ref{tab:tasks}) & 1365.54 & 10.57 & 62.8 \\
    \addlinespace[2pt]
    \bottomrule
  \end{tabular}
  \caption{Seed variation on the learned Llama-2-7B base at $25\%$ utilization
  ($B=672$). Metrics follow Table~\ref{tab:tasks}.}
  \label{tab:seeds}
\end{table}

While this table only presents a small number of seeds run on a single model with a single allocation approach, it still shows that between runs down-stream accuracy can differ significantly.  We therefore match initialization seeds across the uniform, static, and dynamic
arms in Table~\ref{tab:tasks} to control initialization differences within each
comparison.

\section{Additional Layer Utilization Heterogeneity Experiments and Information}
\label{app:heterogeneity}

\subsection{Detailed Experiment Information}
Figure~\ref{fig:heterogeneity} uses learned Llama-2-7B conversion at $25\%$
target utilization ($B=672$), seed \textit{B}, and $3{,}500$ steps
($458{,}752{,}000$ tokens), differing only in allocation.
Panel (a) evaluates WikiText-2 test perplexity every $10$M tokens, giving $46$ evaluations
after initialization. Panel (b) applies a centered $61$-step rolling median
to per-step mean counts. Panel (c) shows $16$ final-step schedules.
Between steps $2{,}501$--$3{,}000$ and $3{,}001$--$3{,}500$, no layer's
mean count changes by more than $0.63$ experts.

\subsection{Per-Layer Profiles Across Models}
\label{app:profiles}
\begin{figure}[!htb]
  \centering
  \includegraphics[width=\linewidth]{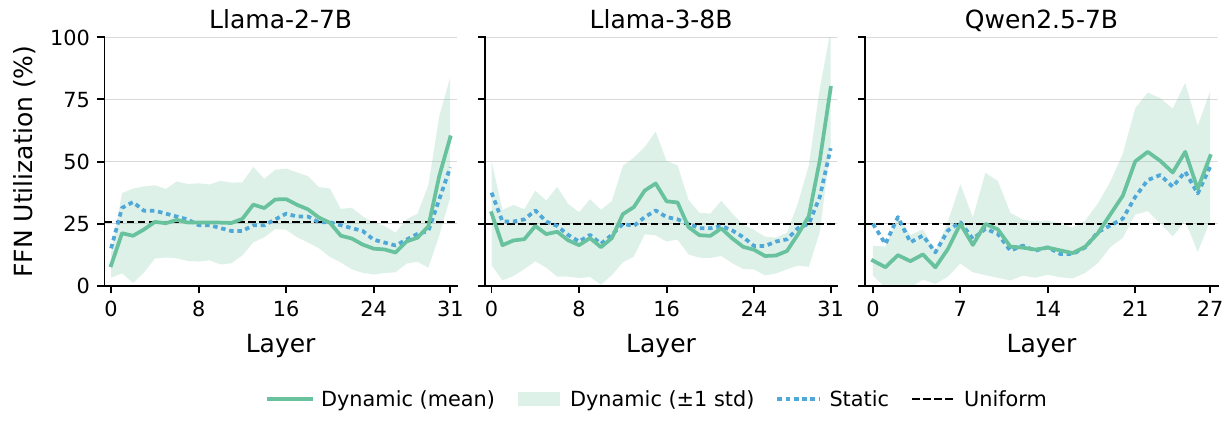}
  \caption{Learned per-layer FFN utilization at $25\%$ utilization on the three main-body
  models.}
  \label{fig:profiles}
\end{figure}

Despite the differences in methodology, static and dynamic allocation both assign more capacity to late layers on all three models (Figure~\ref{fig:profiles}).  In this figure \textit{Static} is the final schedule of the static run in Table~\ref{tab:tasks} and \textit{Dynamic} is the per-layer mean over the $4{,}096$ tokens of the final batch.

\section{Inclusion Temperature Selection}
\label{app:temp}

Table~\ref{tab:temp} varies inclusion temperature in matched static
Llama-2-7B runs initialized with $10$ routed experts per layer.  This data shows the temperature value can have a significant effect on the overall perplexity.  This, in turn, means that selecting a suitable temperature value is imperative to the overall performance of the algorithm. 

\begin{table}[!htb]
  \centering
  \small
  \setlength{\tabcolsep}{6pt}
  \begin{tabular}{@{}lrrrr@{}}
    \toprule
    Inclusion temperature &  Final PPL \\
    \midrule
    Uniform control & 16.30 \\
    $T=0.25$ & 16.57 \\
    $T=1$ & 16.15 \\
    $T=1.5$ & 15.89 \\
    \bottomrule
  \end{tabular}
  \caption{Inclusion temperature for learned static allocation on Llama-2-7B at $\bar{E}=10$
  ($B=320$).}
  \label{tab:temp}
\end{table}

\paragraph{Temperature Selection.}
For an expert of rank $r=\mathrm{rank}_l(t,e)$, Equation~\ref{eq:softincl}
defines the soft inclusion as
\begin{equation*}
  \tilde{R}^{(l)}_{te}
  =\sigma\!\left(\frac{\epsilon_l-r-\tfrac12}{T}\right).
\end{equation*}
Its sensitivity to $\epsilon_l$ peaks at the selection boundary
$r=\epsilon_l-\tfrac12$ and decreases with distance from this boundary.

To illustrate how $T$ controls the range of sensitive ranks, consider
soft inclusions between $0.05$ and $0.95$.
Since $\sigma(-\log 19)=0.05$ and $\sigma(\log 19)=0.95$, we have
\begin{equation*}
  0.05\le\tilde{R}^{(l)}_{te}\le0.95
  \quad\Longleftrightarrow\quad
  \left|r-(\epsilon_l-\tfrac12)\right|\le T\log 19.
\end{equation*}
This interval extends $T\log 19$ ranks on either side of the selection
boundary.
For the initialization $\epsilon_l=10$ in Table~\ref{tab:temp},
the boundary lies at rank $9.5$.
At $T=1.5$, $T\log 19\approx4.42$, so the interval includes ranks
$6$ through $13$.
At $T=0.25$, $T\log 19\approx0.74$, leaving only ranks $9$ and $10$
within the interval.  The interval moves with $\epsilon_l$, so ranks outside it at
initialization can enter the transition region during training.
We select $T$ empirically, with $T=1.5$ giving the lowest perplexity
among the values tested.  There may be superior values for $T$, however, we leave searching for such values for future work.

\end{document}